\documentclass[letterpaper]{article} 
\usepackage{aaai2026}  
\usepackage{times}  
\usepackage{helvet}  
\usepackage{courier}  
\usepackage[hyphens]{url}  
\usepackage{graphicx} 
\usepackage{lipsum}
\usepackage{amsmath}
\usepackage{amssymb}
\usepackage{natbib}  
\usepackage{caption} 
\usepackage{multicol}
\usepackage{multirow}
\usepackage{booktabs}
\usepackage{algorithm}
\usepackage{algorithmic}
\usepackage{enumitem}
\usepackage[most]{tcolorbox}

\usepackage{newfloat}
\usepackage{listings}
\DeclareCaptionStyle{ruled}{labelfont=normalfont,labelsep=colon,strut=off} 
\floatstyle{ruled}
\newfloat{listing}{tb}{lst}{}
\floatname{listing}{Listing}
\definecolor{annocolor}{RGB}{133,112,255}

\title{UniAPO: Unified Multimodal Automated Prompt Optimization}

\author{
    Qipeng Zhu\textsuperscript{\rm 1,2}\footnote{All authors marked with $^\star$ are co-first authors.}, Yanzhe Chen\textsuperscript{\rm 1,3}$^\star$, Huasong Zhong\textsuperscript{\rm 1}$^\star$\footnote{Project Leader.}
    Yan Li\textsuperscript{\rm 1},\\ Jie Chen\textsuperscript{\rm 2}, Zhixin Zhang\textsuperscript{\rm 1},
    Junping Zhang\textsuperscript{\rm 2}, Zhenheng Yang\textsuperscript{\rm 1}\footnote{Corresponding Author}
}
\affiliations{
    \textsuperscript{\rm 1}ByteDance\\
    \textsuperscript{\rm 2}Fudan University\\
    \textsuperscript{\rm 3}National University of Singapore\\

}

\usepackage{bibentry}
\usepackage{subcaption}
\begin{document}

\maketitle

\begin{abstract}


Prompting is fundamental to unlocking the full potential of large language models. To automate and enhance this process, automatic prompt optimization (APO) has been developed, demonstrating effectiveness primarily in text-only input scenarios. However, extending existing APO methods to multimodal tasks—such as video-language generation—introduces two core challenges:
\textit{(\romannumeral1) visual token inflation}, where long visual-token sequences restrict context capacity and result in insufficient feedback signals; 
\textit{(\romannumeral2) a lack of process-level supervision}, as existing methods focus on outcome-level supervision and overlook intermediate supervision, limiting prompt optimization.
We present \textbf{UniAPO}: \textbf{Uni}fied Multimodal \textbf{A}utomated \textbf{P}rompt \textbf{O}ptimization, the first framework tailored for multimodal APO. 
UniAPO adopts an EM-inspired optimization process that decouples feedback modeling and prompt refinement, making the optimization more stable and goal-driven. 
To further address the aforementioned challenges, we introduce a short-long term memory mechanism: historical feedback mitigates context limitations, while historical prompts provide directional guidance for effective prompt optimization.
UniAPO achieves consistent gains across text, image, and video benchmarks, establishing a unified framework for efficient and transferable prompt optimization.

\end{abstract}    
\section{Introduction}
\begin{figure*}[t]
\centering
\includegraphics[width=\linewidth]{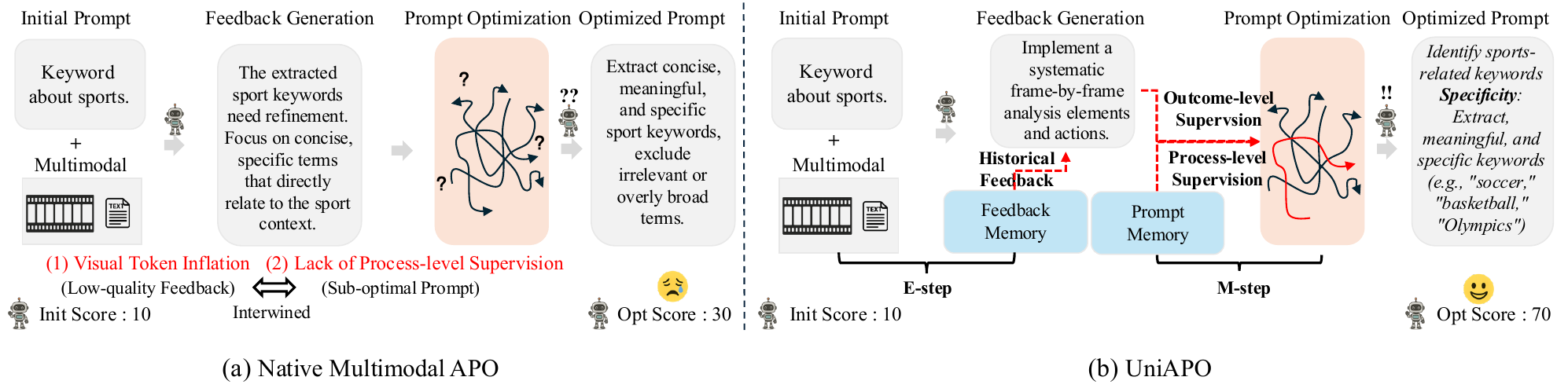} 
\caption{\textbf{Motivation Illustration}: (a) Naively extending text-based APO to multimodal inputs introduces \textit{visual token inflation} and \textit{a lack of process-level supervision}. (b) Our proposal adopts an EM-inspired optimization scheme to iteratively update feedback and prompt memory to solve the above problems.}
\label{fig:qiancai}
\end{figure*}

Recent advances in \textit{automatic prompt optimization} (\textbf{APO}) have enabled large language models to generate and refine prompts without human intervention~\cite{cui2025automatic, li2025survey, ramnath2025systematic}. These methods—ranging from search-based strategies~\cite{APE, Promptbreeder} to feedback-driven approaches~\cite{APO, GPO}—have shown promising results across various natural language tasks~\cite{spiess2025autopdl, saleem2025evolution}. 
Nevertheless, existing methods are largely restricted to unimodal text settings, limiting their applicability in real-world scenarios involving multimodal inputs.
As multimodal large language models become increasingly capable and widely deployed~\cite{zhang2024mm, song2025bridge, chen2025unicode}, there is a growing need for a unified APO framework that can operate seamlessly across text, image, and video inputs.

Extending feedback-driven APO from text to multimodal inputs—by naively appending image or video tokens to existing frameworks—may seem straightforward but quickly encounters two fundamental challenges (shown in Figure~\ref{fig:qiancai}(a)).
First, \textit{visual token inflation}: a single high-resolution image or short video generates hundreds to thousands of tokens~\cite{cao2023efficient, lee2024video},
thereby restricting the number of samples that can be accommodated and resulting in insufficient feedback signals.
Second, \textit{a lack of process-level supervision}: multimodal tasks are inherently more complex~\cite{zhou2025opening, zhang2024worldqa} and demand richer supervision signals to effectively optimize prompts. Relying solely on outcome-level supervision (current feedback) is insufficient, often leading to unstable and suboptimal prompt. And the problems caused by these two challenges will also be intertwined with each other.

These challenges call for rethinking Multimodal APO as \textit{disentangled optimization, expanded feedback signals, and dual-level supervision} (shown in Figure~\ref{fig:qiancai}(b)).
(\romannumeral1) 
The intertwined problems of insufficient feedback signals and 
sub-optimal prompt
create a vicious cycle in multimodal prompt optimization. To break this cycle, we propose a framework inspired by the Expectation-Maximization (EM) algorithm that decouples these problems. 
(\romannumeral2)Visual token inflation quickly saturates limited context, necessitating a long-short term memory mechanism to preserve historical feedback and extend the optimization horizon.
(\romannumeral3) Inspired by reinforcement learning~\cite{yao2023tree,rafailov2023direct}, we argue that supplementing outcome-level supervision with process-level supervision is crucial. This dual-supervision approach stabilizes the optimization toward more performant and robust solutions.

We instantiate these insights in \textbf{UniAPO} (\textbf{Uni}fied multimodal \textbf{A}utomated \textbf{P}rompt \textbf{O}ptimization), the first unified framework adopting an EM-inspired optimization scheme that explicitly decouples feedback modeling from prompt refinement.
In the E-step, UniAPO aggregates valid and diverse feedback using both current errors and semantically relevant historical feedback, ensuring that optimization is informed by a broader context. 
In the M-step, it generates new prompts by integrating short-term candidates with high-quality historical prompts from long-term memory, effectively anchoring the optimization.
These components enable UniAPO to scale to complex multimodal tasks and achieve robust, interpretable prompt optimization.

Our contributions are summarized as follows:
\begin{itemize}
    \item We propose \textbf{UniAPO}, the first unified multimodal APO framework that scales across text, image, and video tasks within a single architecture, achieving state-of-the-art performance compared to existing baselines.
    \item We introduce an \textbf{EM-inspired optimization scheme} that decouples feedback modeling and prompt refinement, yielding a stable optimization process.
    \item We design a \textbf{long-short term memory mechanism} that alleviates \textit{visual token inflation} and \textit{lack of process-level supervision} via historical feedback signals and dual-level supervision.
\end{itemize}

\section{Related Work}
\subsection{Prompt Engineering for MLLMs}


Prompt engineering plays a pivotal role in enabling MLLMs to perform both general reasoning and domain-specific tasks~\cite{chen2023unleashing, mohanty2025future}. A prominent line of research centers on chain-of-thought (CoT) prompting~\cite{cot,mmcot,vcot}, where prompts like ``Think step by step" are used to elicit structured reasoning, especially in spatial contexts. Related works extend this to single-turn reasoning~\cite{shotreason2,shotreason3,shotreason4}, often prompting MLLMs to generate intermediate queries or reflections to enhance interpretability and problem-solving ability.
Beyond reasoning, studies have explored prompt formatting~\cite{format1,format2,format3} as a way to improve response consistency, especially in scenarios requiring tool use, layout understanding, or constrained output forms.
To address task-specific needs, researchers have developed domain-adapted prompts across a wide range of applications. This includes open-vocabulary grounding~\cite{grounding1,grounding2,grounding3}, semantic segmentation~\cite{seg1,seg2}, and visual question answering (VQA)~\cite{vqa,vqa1}, where prompt designs are often tailored to the data modality and task structure.
Despite promising results, these approaches rely heavily on manual prompt design, which becomes increasingly infeasible as MLLMs are deployed across more complex, diverse, and open-ended domains. This limitation has spurred growing interest in automated prompt optimization techniques~\cite{Aflow}, aiming to scale prompt engineering in a systematic and adaptive manner.

\subsection{Automatic Prompt Optimization (APO)}

APO aims to automatically discover effective prompts for LLMs and MLLMs, reducing manual effort while enhancing generalization across diverse tasks~\cite{cui2025automatic, qu2025proapo, ramnath2025systematic, do2025makes}. Existing approaches fall into two main paradigms: search-based optimization and feedback-driven refinement.
Search-based methods explore the prompt space by iteratively sampling and evaluating candidates~\cite{davari2025rethinking, zhang2024neural}. APE~\cite{APE} frames prompt construction as a discrete optimization task, with LLMs generating and scoring prompts in a closed loop. Subsequent works adopt evolutionary strategies~\cite{prompyevol, Promptbreeder} or treat LLMs as black-box optimizers~\cite{prompt_opt}. However, these methods often suffer from search path explosion in semantically complex or open-ended settings, limiting their scalability in multimodal domains.
Feedback-driven methods improve stability by introducing an intermediate phase: models analyze failure cases and generate textual feedback, which is then used to revise prompts~\cite{agarwal2025promptwizard}. APO~\cite{APO} pioneered this paradigm, viewing feedback as a textual “gradient” to guide optimization. Later work extends this idea with analogical reasoning~\cite{GPO}, pseudo-gradient propagation~\cite{Textgrad}, memory-augmented reflection~\cite{ERM}, and strategic self-guidance~\cite{strago}, achieving strong performance in text-only tasks.
Despite success in text tasks, feedback-based APO struggles in multimodal contexts: visual token inflation and lack of process-level supervision. We alleviates visual token inflation and lack of process-level
supervision via historical feedback signals and dual-level
supervision by designing a long-short term memory mechanism.

\begin{figure*}[t]
\centering
\includegraphics[width=\linewidth]{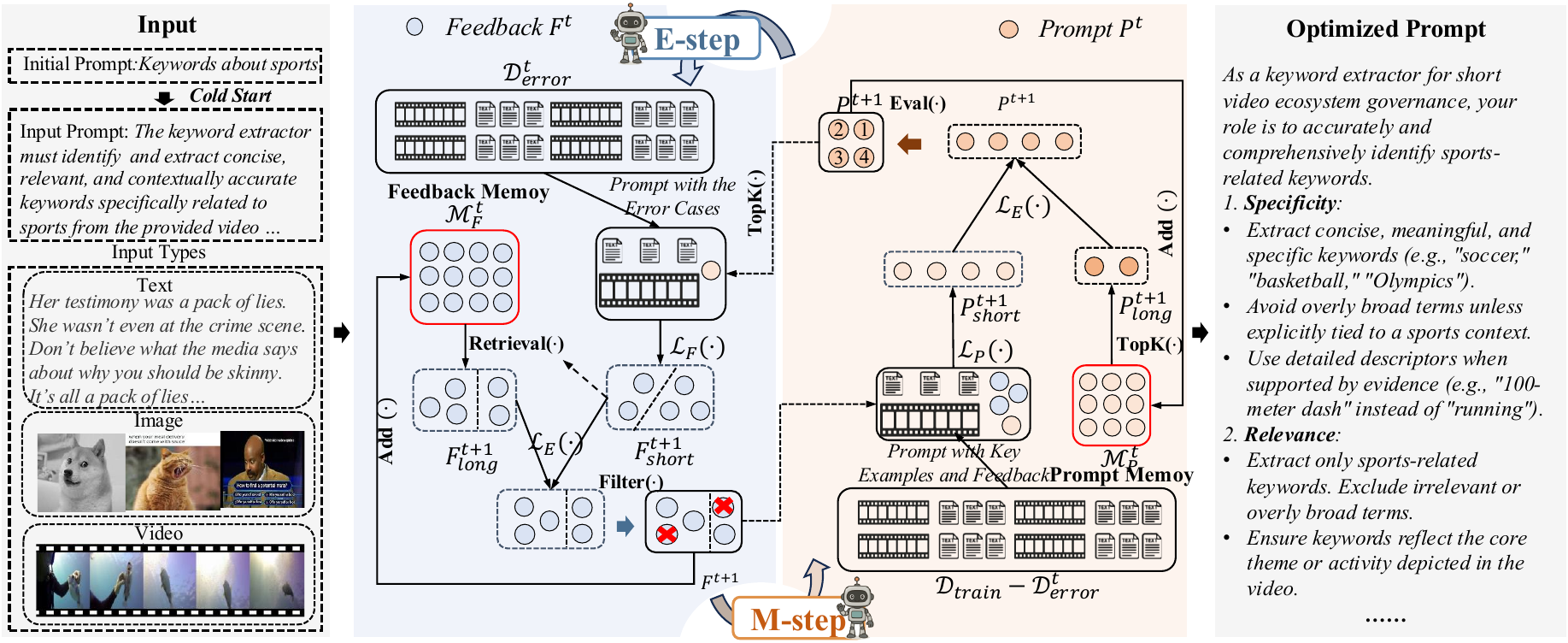} 
\caption{ Illustration of our UniAPO framework for UniAPO. Starting with a simple prompt initialized by an MLLM (left), UniAPO iteratively refines it into a structurxed and knowledgeable prompt (right) using an Expectation-Maximization (EM) algorithm. The E-step generates long- and short-term feedback from the current prompt, which is then used in the M-step to update the prompt, enabling optimization across diverse data types.}
\label{fig:method}
\end{figure*}
\section*{Preliminaries}
\label{sec:preliminaries}

\subsection{Problem Formulation and Baseline}
\label{sec:formulation}
Let the datasets be denoted as $\mathcal{D}_{\text{train}}$, $\mathcal{D}_{\text{dev}}$, and $\mathcal{D}_{\text{test}}$, each consisting of sample-label pairs $(x, y)$.
We consider a system of frozen MLLMs with different system prompts as alternates roles: a task model $\mathcal{L}_{T}$ for prediction, a feedback model $\mathcal{L}_{F}$ for generating feedback, a prompt optimization model $\mathcal{L}_{P}$, and an evolution model $\mathcal{L}_{E}$.
Details of system prompts are stated in the Appendix.
Our primary objective is to find the optimal prompt $P^*$ that maximizes the expected performance on a given dataset $\mathcal{D}_\text{test}$:
\begin{equation}
\label{eq:objective}
P^* = \underset{P \in \mathcal{P}}{\text{argmax}} \, \mathbb{E}_{(x,y) \in \mathcal{D}_\text{test}}[\text{Eval}(\mathcal{L}_{T}(x; P), y)],
\end{equation}
where $\mathcal{P}$ represents the space of all possible prompts and  $\text{Eval}(\cdot)$ is the evaluation metric.

Then we establish a baseline method based on feedback-driven Automatic Prompt Optimization (APO).
In a naive multimodal feedback-driven APO~\cite{APO} loop, the optimization process is iterative. 
At each step $t$, we identify an error set $\mathcal{D}^t_{\text{error}} \subseteq \mathcal{D}_{\text{train}}$ where the task model $\mathcal{L}_{T}$ fails with the current prompt $P^t$. 
Subsequently, the feedback model $\mathcal{L}_{F}$ generates feedback $F^{t+1}$ based on $\mathcal{D}^t_{\text{error}}$ and $P^t$. 
Finally, the prompt optimization model $\mathcal{L}_{P}$ optimizes the prompt $P^t$ using the feedback $F^{t+1}$ to produce an improved prompt $P^{t+1}$. 
However, this straightforward feedback-driven approach encounters two significant challenges. Details of system prompts are stated in the Appendix.

\subsection{Core Challenges}
\label{sec:challenges}

A naive multimodal APO framework faces two critical, intertwined challenges: visual token inflation~\cite{cao2023efficient, lee2024video} and a lack of process-level supervision~\cite{PRM}. Visual token inflation stems from the feedback generator's ($L_F$) finite context, which yields low-quality feedback by failing to process all historical and current errors. Concurrently, the prompt optimizer ($L_P$) receives only this outcome-level supervision, leading to sub-optimal prompts. These issues create a vicious cycle of mutual degradation, making a simultaneous solution exceptionally difficult.


\section{Methodology}
To tackle the two intertwined challenges of Visual Token Inflation and a Lack of Process-level Supervision, we propose a novel framework named \textbf{Uni}fied Multimodal \textbf{A}utomatic \textbf{P}rompt 
\textbf{O}ptimization (\textbf{UniAPO}). Our approach is inspired by the Expectation-Maximization (EM) algorithm and employs a divide-and-conquer strategy to decouple the problem, as illustrated in Figure~\ref{fig:method}. 
UniAPO consists of two main steps: an E-step designed to address Visual Token Inflation, and an M-step to counter a Lack of Process-level Supervision. This design effectively breaks the vicious cycle arising from the interplay of these two challenges.

\subsection{Overall Architecture}
\label{sec:overview}
A core component of UniAPO is the integration of memory to leverage historical information. We introduce a feedback memory, $\mathcal{M}_F^t$, and a prompt memory, $\mathcal{M}_P^t$, to store all generated feedback and prompts up to iteration $t$. 

Specifically, our method begins with a simple phase. We use the prompt optimization model, $\mathcal{L}_P$, to refine a simple, sample-agnostic initial prompt (e.g., ``keywords about sports'') to obtain a superior input prompt, $P^0$. This ensures that the optimization process starts from a more reasonable point in the optimization space. The optimization then proceeds iteratively through the E-step and M-step.

\noindent\textbf{E-Step:} 
At iteration $t$, the current prompt $P^t$ is used with the multimodal inputs to perform inference (assised by $\mathcal{L}_T$), resulting in an error set $\mathcal{D}_{\text{error}}^{t}$. This error set, along with the feedback memory $\mathcal{M}_F^t$, is then processed by the feedback model $\mathcal{L}_F$ (potentially assisted by a evolution model $\mathcal{L}_E$) to generate new, targeted feedback $F^{t+1}$. The feedback memory is subsequently updated with this new information. The entire process can be expressed as:
\begin{equation}
\label{eq:estep}
(F^{t+1}, \mathcal{M}_F^{t+1}) = \text{E-Step}(\mathcal{D}_{\text{error}}^{t}, \mathcal{M}_F^t; \mathcal{L}_F, \mathcal{L}_E).
\end{equation}

\noindent\textbf{M-Step:} 
In the subsequent M-step, the newly generated feedback $F^{t+1}$ and the prompt memory $\mathcal{M}_P^t$ are used to guide the prompt optimization model $\mathcal{L}_P$ (also assisted by $\mathcal{L}_E$). This step refines the current prompt $P^t$ to produce an improved prompt $P^{t+1}$ for the next iteration, and the prompt memory is updated accordingly. This step can be formulated as:
\begin{equation}
\label{eq:mstep}
(P^{t+1}, \mathcal{M}_P^{t+1}) = \text{M-Step}(F^{t+1}, \mathcal{M}_P^t, P^t; \mathcal{L}_P, \mathcal{L}_E).
\end{equation}

In the following subsections, we will elaborate on how the E-step and M-step are specifically designed to address the challenges of Visual Token Inflation and a Lack of Process-level Supervision, respectively.

\subsection{E-step: Multimodal Feedback Generation}
The E-step is specifically designed to combat the Visual Token Inflation challenge during the feedback generation phase. The essence of this problem lies in a practical constraint: the feedback model, $\mathcal{L}_F$, has a finite context window. As the generation process iterates, the cumulative set of all encountered errors can easily grow to exceed this capacity. Consequently, at iteration $t$, it becomes infeasible to feed the entire raw error history into $\mathcal{L}_F$ for consideration.

To overcome this limitation, we introduce a short- and long-term memory mechanism. Our key insight is that the complete error history can be effectively represented by two distinct components:
\begin{itemize}
    \item \textbf{Short-term Information}: The current error set, $\mathcal{D}_{\text{error}}^t$, which captures the model's most recent failures and is used by $\mathcal{L}_P$ to generate the next feedback, $F^{t+1}$.
    \item \textbf{Long-term Information}: The feedback memory, $\mathcal{M}_F^t$, which stores a cumulative history of past errors and their associated corrective feedback.
\end{itemize}
The E-step is to first extract information from these two sources and then unify them, ensuring that a holistic view of all errors can be processed within the limited context of $\mathcal{L}_F$. 

\subsubsection{Short-Term Feedback Generation.}
A practical challenge remains: even the most recent error set, $\mathcal{D}_{\text{error}}^t$, can be too large to fit into the context window of $\mathcal{L}_F$ in a single pass. To manage this, we adopt a hierarchical strategy inspired by techniques in multimodal Retrieval-Augmented Generation (RAG)~\cite{visrag}. The procedure first clusters $\mathcal{D}_{\text{error}}^t$ to group semantically similar failures, enabling $\mathcal{L}_F$ to produce more stable feedback on common error patterns. Subsequently, to adhere to the model's context limit, each resulting cluster is processed in smaller \textit{chunks}. Feedback is generated for each chunk and then aggregated to represent the entire cluster's error profile, as depicted in Figure~\ref{fig:method}.
The entire process of generating the short-term feedback, denoted as $F^{t+1}_{\text{short}}$, can be formally expressed as:
\begin{equation}
\begin{array}{cc}
F^{t+1}_{\text{short}}     & =\mathcal{L}_{\mathcal{F}}(P_t, \text{Clustering}({D}^{t}_\text{error})),

\end{array}
\end{equation}
where $\text{Clustering}(\cdot)$ is the DBSCAN algorithm using BGE-m3~\cite{bge} embeddings.




\subsubsection{Long-Term Feedback Generation.}
A naive inclusion of the entire memory $\mathcal{M}_F^t$ is suboptimal, as obsolete feedback for corrected errors can introduce semantic noise. To address this, we shift from simple summarization to targeted retrieval. Specifically, we use the newly generated short-term feedback, $F^{t+1}_{\text{short}}$, as a dynamic query. The feedback derived from each error cluster acts as a separate query to retrieve the most relevant entries from the memory $\mathcal{M}_F^t$. These retrieved historical records are then aggregated to form a potent and contextually relevant long-term feedback, $F^{t+1}_{\text{long}}$, as illustrated in Figure~\ref{fig:method}, where $\text{Retrieval}(\cdot,\cdot)$ denotes the retrieval process. The entire generation process can be formulated as:
\begin{equation}
\label{eq:long_term_feedback}
F^{t+1}_{\text{long}}=\text{Retrieval}(F^{t+1}_{\text{short}},\mathcal{M}_F^t).
\end{equation}
\subsubsection{Short- and Long-Term Feedback Evolving}
To combine the short-term ($F^{t+1}_{\text{short}}$) and long-term ($F^{t+1}_{\text{long}}$) feedback, we devise a two-step process. First, inspired by evolutionary algorithms~\cite{evol}, an ``Evolver'' MLLM, $\mathcal{L}_E$, fuses the two streams, guided by a system prompt to resolve conflicts and merge salient information. Second, to guarantee utility, the resulting candidate feedback undergoes a filtering step, $\text{Filter}(\cdot)$, inspired by ERM~\cite{ERM}. This step validates the feedback by retaining only suggestions that demonstrably correct errors in the original set $\mathcal{D}_{\text{error}}^t$. The generation of the final, validated feedback $F^{t+1}$ is formulated as:
\begin{equation}
\label{eq:final_feedback}
F^{t+1}   =\text{Filter}(\mathcal{L}_E(F^{t+1}_{\text{short}},F^{t+1}_{\text{long}}),\mathcal{D}_{\text{error}}^t,P^t;\mathcal{L}_T)
\end{equation}
where $F^{t+1}$ is added into $\mathcal{M}_F^t$ to gain $\mathcal{M}_F^{t+1}$ as depicted in Equation~\eqref{eq:fm_add}:
\begin{equation}
\label{eq:fm_add}
\mathcal{M}_F^{t+1}= \text{Add}(\mathcal{M}_F^{t},F^{t+1}).
\end{equation}

\subsection{M-step: Multi-modal Prompt Optimization}
The M-step resolves the outcome-only supervision problem by synergizing two distinct supervisory signals for prompt optimization.
\begin{itemize}
    \item Outcome-level Supervision: Following naitive feedback-driven methods~\cite{APO}, we use the immediate feedback, $F^{t+1}$, to perform a tactical update on the current prompt, $P^t$, yielding a short-term prompt, $P^{t}_\text{short}$.
    \item Process-level Supervision: Inspired by PRMs~\cite{PRM}, we introduce a novel process-level signal by distilling a \textit{long-term prompt} from the entire prompt history, $\mathcal{M}_P^t$. This prompt embodies stable, historically effective strategies.
\end{itemize}
The final prompt, $P^{t+1}$, is synthesized by modulating the short-term prompt with the strategic guidance from the long-term prompt. This ensures that our updates are not only responsive to immediate failures but are also grounded in a history of successful optimizations, leading to superior robustness and performance.


\subsubsection{Short-Term Prompt Optimization.}
Our process begins with generating a Short-Term Prompt, $P^{t+1}_\text{short}$, by leveraging an MLLM optimizer, $\mathcal{L}_P$, to refine the current prompt $P^t$. This refinement is guided by the recent, coarse-grained feedback $F^{t+1}$. To ensure the optimizer maintains a robust understanding of the task~\cite{ICL}, we also provide it with a set of positive examples, $\text{Sample}(\cdot)$, sampled from $\mathcal{D}_\text{train}-\mathcal{D}^{t}_\text{error}$. This prevents over-fitting to recent failures and is formally expressed as:
\begin{equation}
    P^{t+1}_\text{short}=\mathcal{L}_P(P_t,F^{t+1},\text{Sample}\left(\mathcal{D}_\text{train}-\mathcal{D}^{t}_\text{error})\right)
\end{equation}
We run the optimizer $\mathcal{L}_P$ multiple times to generate a diverse set of candidate prompts, as shown in Figure~\ref{fig:method}.

\subsubsection{Long-Term Prompt Generation}
To ensure that our process supervision signal is derived from high-quality prompts, we filter the prompt history rather than using it wholesale. We recognize that underperforming prompts can provide misleading guidance. Therefore, we select only the top-$k$ historical prompts from $\mathcal{M}_P^t$ based on their scores on the $\mathcal{D}_\text{dev}$. This selection is performed via a Top-K algorithm, yielding $P^{t+1}_\text{long}$:
\begin{equation}
    P^{t+1}_\text{long} = \text{TopK}(\mathcal{M}_P^t, k)
    \label{eq:topk_selection}
\end{equation}

\subsubsection{Short- and Long-term Prompt Evolving.}
To effectively fuse the process and outcome signals, we introduce a step inspired by evolutionary crossover. We task the MLLM optimizer, $\mathcal{L}_E$, to act as a supervisor that intelligently synthesizes the short-term prompt with the wisdom from the long-term prompts. This supervised crossover allows the current prompt to adopt the proven advantages of its predecessors in a structured way. The process is defined as:
\begin{equation}
    P^{t+1} = \mathcal{L}_E(P^{t+1}_\text{short}, P^{t+1}_\text{long})
    \label{eq:guided_synthesis}
\end{equation}
The generated prompt $P^{t+1}$ is first evaluated on $\mathcal{D}_\text{dev}$, and its score is recorded as it is integrated into the prompt memory, which is updated to $\mathcal{M}_P^{t+1}$:
\begin{equation}
\mathcal{M}_P^{t+1}= \text{Add}(\mathcal{M}_P^{t},P^{t+1}).
\end{equation}

To prevent premature convergence and expand the optimization horizon, we then employ a beam search mechanism. 
Specifically, we select the top-$b$ prompts from $\mathcal{M}_P^{t+1}$ based on their scores. These $b$ prompts become parallel `beams' for the next iteration.

\section{Experiment}

\setlength{\tabcolsep}{2.5pt}
\begin{table*}[!ht]
\small
\centering
\begin{tabular}{cccccccccccc}
\toprule
&\multicolumn{3}{c}{Text CLS} & \multicolumn{1}{c}{Text GEN}                &    Image CLS                                   & \multicolumn{2}{c}{Video CLS}   & \multicolumn{4}{c}{Video KE}                  \\ \cmidrule{2-12} 
\multirow{-2}{*}{Method}  &  LIAR & BBH & ETHOS & WebNLG         & Meme & Static & Occlusion layer & Beauty        & Sport         & Travel        & Food          \\ \midrule
\multicolumn{12}{c}{\textit{GPT4o as Predictor}}  \\
\midrule
Vanilla    &25.3 & 69.4 & 88.6 & 50.9     &  25.8  & 71.2& 25.6  & 36.7          & 55.8          & 43.5          & 24.6          \\
Vanilla + CoT~\cite{cot}    &56.9 & 90.7 & 95.0 & 51.1 & 25.6     &80.1 & 50.0  & 46.9 & 63.9 &54.1& 31.5\\
EvoPrompt*~\cite{prompyevol}   &58.6&  92.7 &96.6 &  50.5 &        26.9  &       82.8     & 33.3  & 47.4 & 56.2  & 44.9& 24.7\\
ERM*~\cite{ERM}     &65.2&95.4  &95.6&    52.1    &  28.6        &       80.1     &  61.5 & 68.3 & 69.3 & 57.4  & 40.3\\
\textbf{UniAPO}&\textbf{78.7}&\textbf{99.4}&\textbf{98.1}& \textbf{53.2} & \textbf{37.6}  & \textbf{86.3}            & \textbf{70.3}   & \textbf{74.7} & \textbf{78.3} & \textbf{60.9} & \textbf{54.3} \\
\midrule
\multicolumn{12}{c}{\textit{QwenVL2.5-72B as Predictor}}  \\
\midrule
Vanilla &2.0&44.7&89.0&   44.3     &   24.7        & 0.0       & 25.6            & 28.7          & 50.0          & 45.9          & 27.6          \\
Vanilla + CoT~\cite{cot} &49.4&93.2&97.6&46.3&24.6&54.5& 41.9& 43.9 & 58.6&47.1& 25.3\\
EvoPrompt*~\cite{prompyevol}   &50.6&94.1&98.0& 46.3  &  25.8         &     78.2       & 30.0  & 44.3& 52.8 & 46.1&27.8 \\
ERM*~\cite{ERM}  &67.4&93.3&98.2&    52.3   &      28.2     &       59.8     &  63.2 & 64.0 & 64.1 & 51.2 & 41.4\\
\textbf{UniAPO}&\textbf{73.1}&\textbf{95.8}&\textbf{98.9}&\textbf{54.4} &\textbf{35.7}&     \textbf{83.1} &  \textbf{67.9}  & \textbf{75.2} & \textbf{76.8} & \textbf{63.7} & \textbf{48.6}  \\
\bottomrule
\end{tabular}
\caption{Performance comparison using GPT-4o vs. QwenVL2.5-72B as the predictor, optimized by our UniAPO framework. UniAPO's other internal components are implemented using GPT-4o. All experiments are conducted on 11 datasets including text classiifaction (``Text CLS''), text generation (``Text GEN''), image classification (``Image CLS'') and video classification (``Video CLS'') and video keyword extraction (``Video KE'').}
\label{exp:exp_comp_main}
\end{table*}

\subsection{Experimental Setting}


\subsubsection{Datasets.}

We evaluate \textbf{UniAPO} across \textit{text}, \textit{image}, and \textit{video} domains on both classification and generation tasks: (1) \textbf{Text}: LIAR~\cite{wang2017liar} (fake news classification), BBH-navigate~\cite{bbh} (multi-step instruction following), ETHOS~\cite{ethos} (hate speech detection), and WebNLG~\cite{webnlg} (structured-to-text generation). (2) \textbf{Image}: Meme~\cite{meme} (multi-image classification requiring semantic alignment via prompt reasoning). (3) \textbf{Video}: An in-house dataset from an international platform, covering static classification (low-motion detection), occlusion classification (identifying overlays), and open-domain keyword extraction (generating keywords from multimodal metadata) across Beauty, Sport, Travel, and Food themes. More details are stated in Appendix.


\subsubsection{Evaluation Metrics.}

Tasks are grouped by domain with corresponding metrics: \textbf{Text classification} (\textit{LIAR}, \textit{ETHOS}, \textit{BBH-navigate}): binary F1 score; \textbf{Text generation} (\textit{WebNLG}): ROUGE-L; \textbf{Image classification} (\textit{Meme}): multi-class F1-micro; \textbf{Video classification} (\textit{Static}, \textit{Occlusion}): binary F1; \textbf{Multimodal keyword extraction} (video, four themes): F1-score More details are stated in Appendix.


\subsubsection{Baselines.}
For all tasks, we compare UniAPO against standard prompting, Chain-of-Thought (CoT) prompting~\cite{cot}, and two prominent categories of automatic prompt optimization:
(1) Search-based methods (e.g., EvolPrompt~\cite{prompyevol}), which iteratively mutate and select prompts;
(2) Feedback-based methods (e.g., ERM~\cite{ERM}), which update prompts based on performance signals.
\subsubsection{Implementation Details.}
All primary experiments use GPT-4o~\cite{achiam2023gpt} as the underlying MLLM across all stages of the UniAPO pipeline. Prompts are initialized with minimal handcrafted templates, denoted as ``Simple Prompt'' to simulate a low-resource setting.
In additional experiments, we replace GPT-4o with QwenVL2.5-72B~\cite{bai2025qwen2} as the predictor to evaluate cross-model generalization, while keeping the other components unchanged. We also explore settings with more structured initial prompts, as detailed in relevant sections.
\subsection{Comparision Study}


\subsubsection{Comparision with different tasks.}
UniAPO sets a new state-of-the-art across a diverse suite of multimodal tasks as shown in Table~\ref{exp:exp_comp_main}, consistently outperforming existing baselines. Its superior performance and stability, particularly on video tasks, are driven by our unified memory mechanism that combats visual token inflation and a lack of process-level supervision. Underscoring its robustness, UniAPO maintains its effectiveness when the backbone model is switched from GPT-4o to Qwen2.5VL-72B, proving the generalizability of our framework.

\subsubsection{Generalization of UniAPO.}
\begin{figure}[ht]
\centering
\includegraphics[width=\linewidth]{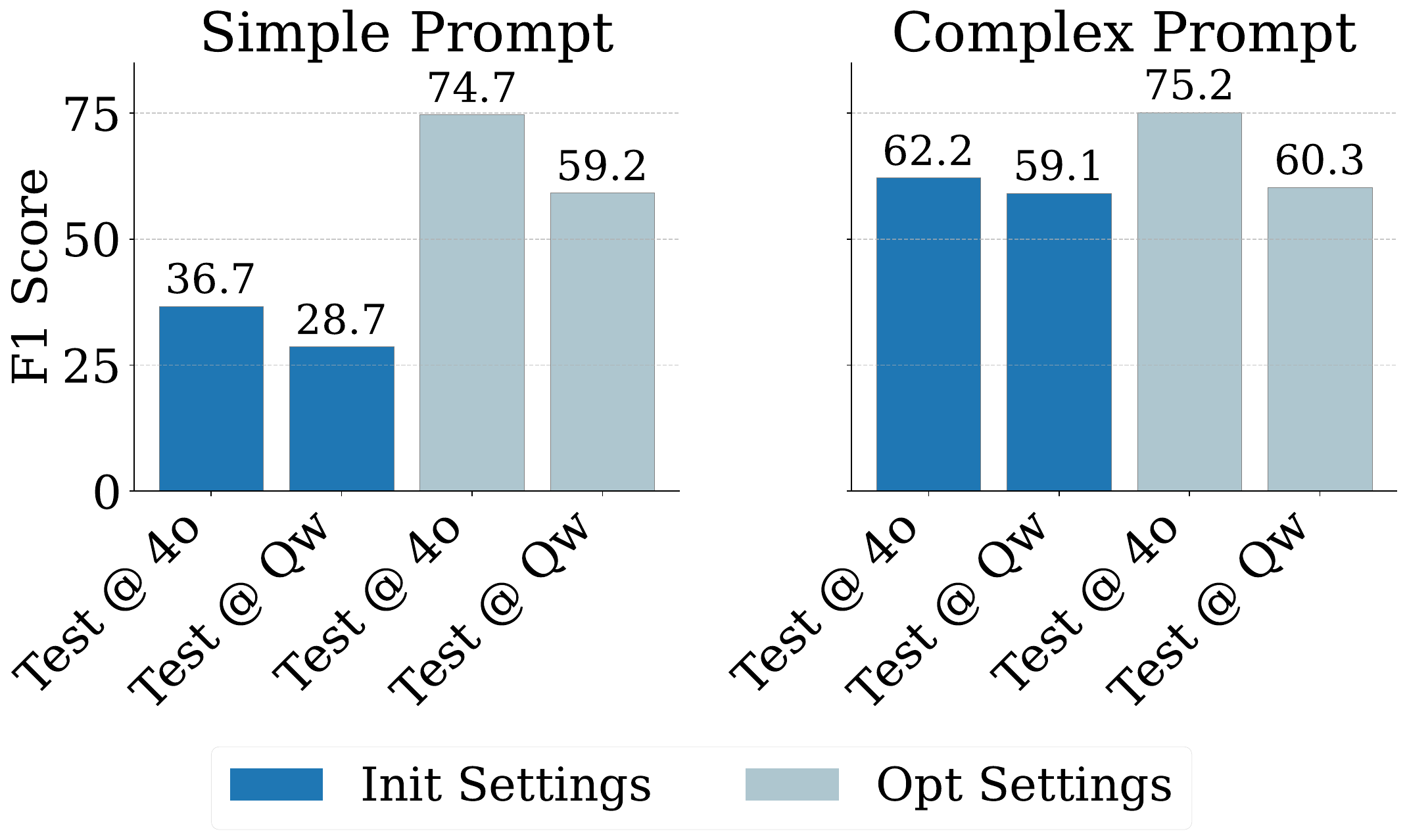} 
\caption{Evaluating the robustness and transferability of UniAPO in beauty keyword extraction. 
The table compares performance from ``Simple'' and ``Complex'' initial (``Init'')  prompts against our optimized prompts (``Opt'') based on GPT4o. We use ``Test @ 4o'' and ``Test @ Qw'' respectively represent the predictor types when testing.}
\label{fig:exp_comp_gen}
\end{figure}

UniAPO demonstrates strong generalization, which we validate through two key experiments: robustness to initialization and cross-model transfer (Figure~\ref{fig:exp_comp_gen}).
\begin{itemize}
    \item Robustness to Initialization: UniAPO is largely insensitive to the quality of the initial prompt. It consistently elevates the performance of both simple and complex starting prompts, as evidenced by the significant gap between ``Opt Settings'' and ``Init Settings'' on ``Test @ 4o''. This robustness is a direct result of its EM framework, which iteratively refines the solution, and its process-level supervision.
    \item Cross-Model Transferability: Prompts optimized by UniAPO  transfer effectively across different architectures. When prompts optimized on GPT-4o are transferred to different the testing predictor settings, such as Qwen2.5-VL-72B, they retain a substantial performance advantage over the original prompts ("Test @ Qw" with "Opt Settings" vs. "Init Settings").
\end{itemize}

\subsubsection{Efficiency of UniAPO.}
\begin{figure}[ht]
\centering
\includegraphics[width=\linewidth]{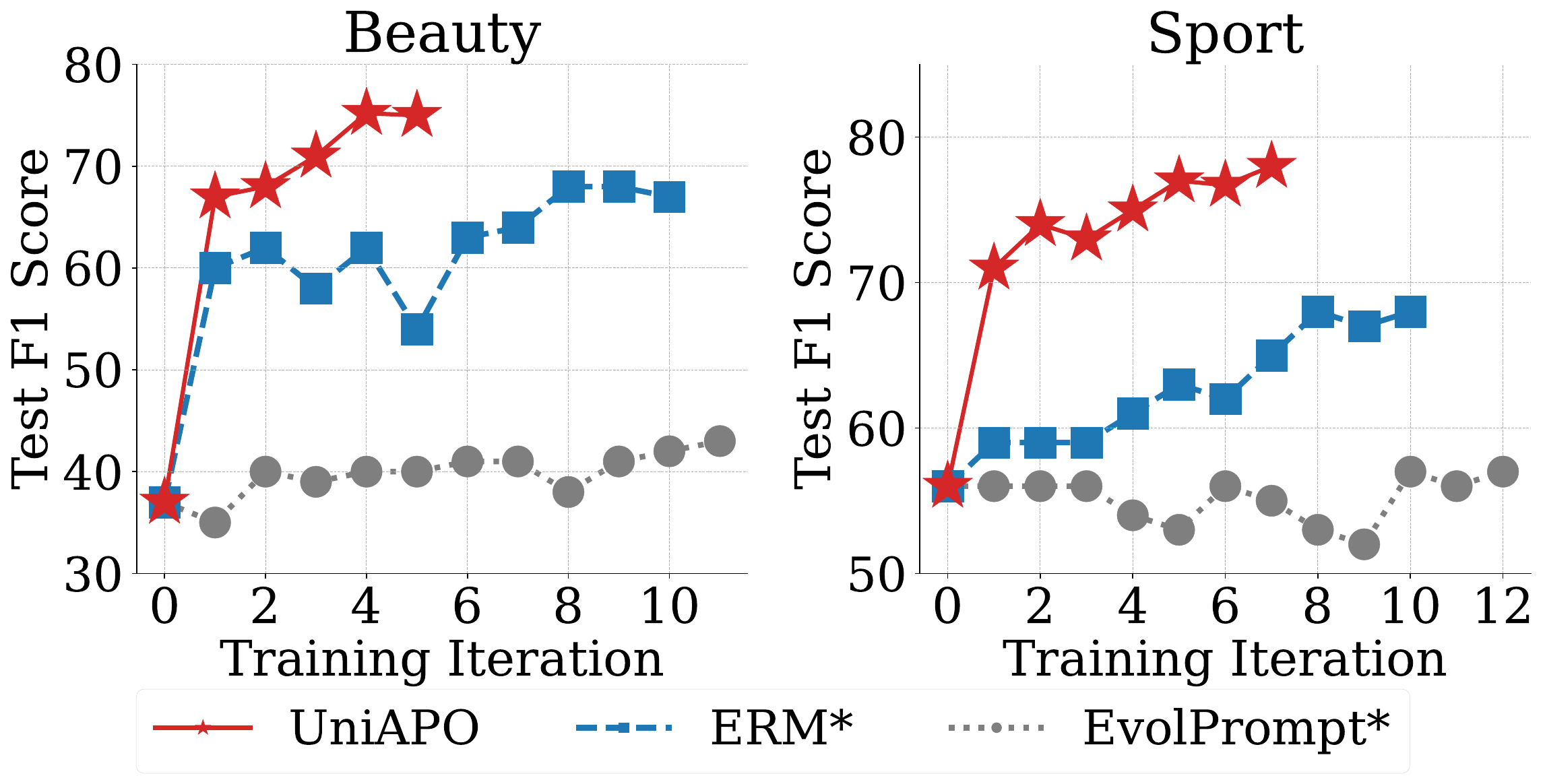} 
\caption{Optimization efficiency and performance comparison. This figure illustrates the Testing F1-score progression for UniAPO, ERM*, and EvolPrompt* over iterations. 
}
\label{fig:exp_comp_efficiency}
\end{figure}

UniAPO is significantly more efficient than baselines, reaching superior performance in fewer optimization steps (Figure~\ref{fig:exp_comp_efficiency}). This is attributed to its EM-inspired framework, which creates a virtuous cycle: an E-step refines feedback by mitigating visual inflation, and an M-step uses dual-level supersion to optimize prompt effectively. This closed-loop process accelerates convergence, demonstrating that UniAPO delivers state-of-the-art results with greater sample and compute efficiency.

\subsection{Analysis Study}

\begin{figure}[ht] 
    \centering 
    \begin{subfigure}[b]{0.23\textwidth}
        \centering
        \includegraphics[width=\linewidth]{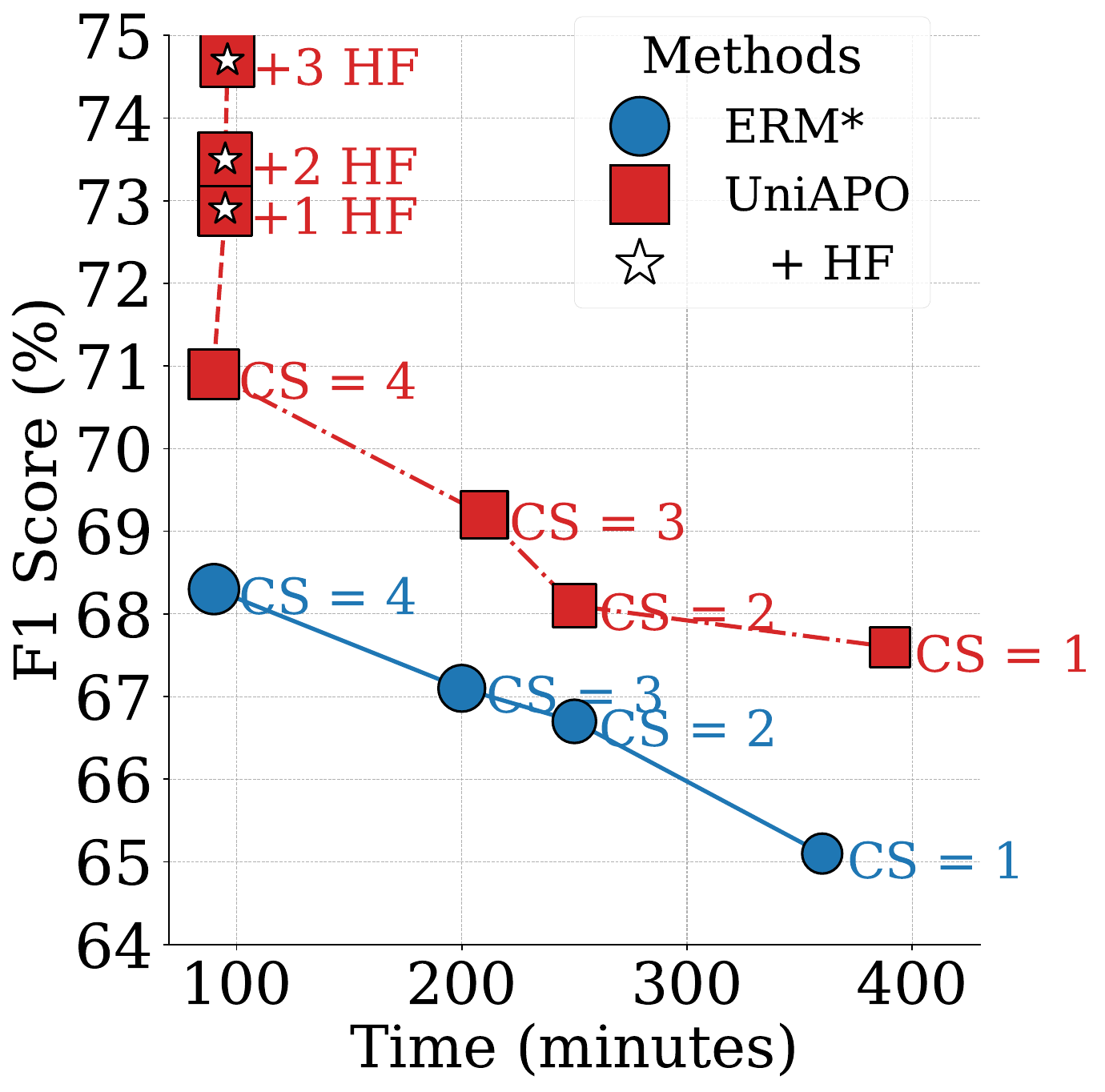} 
        \caption{Effect of increasing chunk sizes (``CS'') and historical feedback (``HF'').}
        \label{fig:exp_ana_vtf}
    \end{subfigure}
    \hfill 
    \begin{subfigure}[b]{0.23\textwidth}
        \centering
        \includegraphics[width=\linewidth]{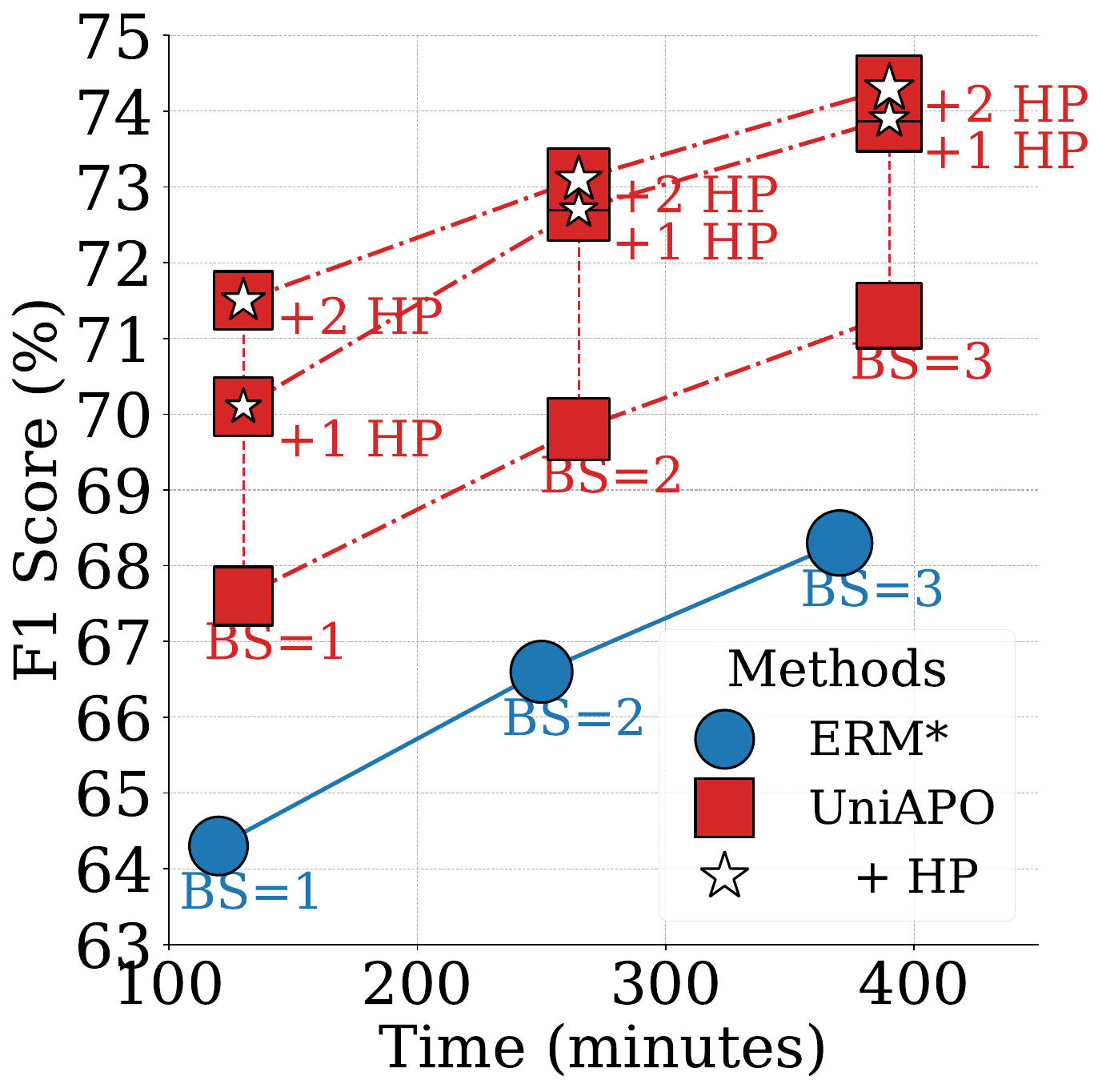}
        \caption{Effect of increasing beam size (``BS'') and historical prompts (``HP'').}
        \label{fig:exp_ana_path}
    \end{subfigure}

    \caption{UniAPO is proven to be both practically efficient and highly effective to alleviate visual token inflation and a lack of process-level supervision.}
    \label{fig:main_figure}
\end{figure}
\subsubsection{Visual Token Inflation.}
Here, we empirically validate the Visual Token Inflation (VLF) bottleneck and the efficacy of our historical feedback solution (Figure~\ref{fig:exp_ana_vtf}). We first establish that while performance scales with the number of input errors, it inevitably saturates as it hits the feedback generator's context limit. This confirms the VLF problem. Critically, introducing our historical feedback at this saturation point yields further, significant performance gains. This result demonstrates that our long-term memory mechanism effectively compensates for the limited context window, enriching the feedback generation process with vital historical information.

\subsubsection{A Lack of Process-level Supervision.}
Figure~\ref{fig:exp_ana_path} validates our core hypothesis: dual-level supervision is essential for robust prompt optimization. We show that a feedback-only baseline (blue line) is insufficient. By augmenting this with process-level supervision from varying numbers of historical prompts, our method consistently boosts performance across all tested beam sizes, critically, with no computational overhead. This demonstrates that integrating process-level guidance with outcome-based feedback is key to achieving stable and superior optimization results.

\subsection{Ablation Study}
\begin{table}[ht]
\small
\centering
\begin{tabular}{ccccc}
\toprule
\multirow{2}{*}{E-step} & \multirow{2}{*}{M-step} & Video CLS & \multicolumn{2}{c}{Video KE} \\ \cmidrule{3-5} 
&    & Occlusion layer   & Beauty    & Sport             \\ \midrule
 &   & 25.6 & 36.7 & 55.8     \\
$\checkmark$      &      &          59.3         &   66.3 &  75.1\\
&    $\checkmark$    &61.2 &    67.8    &  73.0   \\
$\checkmark$  &  $\checkmark$    & \textbf{70.3}  & \textbf{75.2}  & \textbf{78.3}      \\ \bottomrule
\end{tabular}
\caption{Ablation of E-step and M-step.}
\label{tab:ablation_em}
\end{table}
\subsubsection{Ablation of E-step and M-step.}
As shown in Figure~\ref{tab:ablation_em}, our ablation study confirms the synergistic relationship between UniAPO's E-step and M-step. While both prompt optimization (M-step) and feedback generation (E-step) are individually effective, yielding significant gains when used alone, the full framework that alternates between them performs best, which validates that the complementary interaction of these two steps is critical to UniAPO's capabilities.
\setlength{\tabcolsep}{1.5pt}
\begin{table}[ht]
\small
\centering
\begin{tabular}{ccccc}
\toprule
\multirow{2}{*}{FG Type} & \multirow{2}{*}{PO Type} & Video CLS & \multicolumn{2}{c}{Video KE} \\ \cmidrule{3-5}
    &     & Occlusion layer      & Beauty             & Sport             \\ \midrule
ERM*   & ERM*   &61.5 & 68.3 & 69.3                  \\
UniAPO  & ERM*& 65.5 & 73.1        &  74.3                 \\
ERM*& UniAPO    & 65.6 &  70.7   &    76.7               \\
UniAPO                   & UniAPO                   & \textbf{70.3}  & \textbf{75.2}  & \textbf{78.3}                     \\ \bottomrule
\end{tabular}
\caption{Comparison with different combainations between Feedback Generation methods (FG) and Prompt Optimization (PO) methods.}
\label{tab:abl_fg_po}
\end{table}

\subsubsection{Feedback Generators and Prompt Optimizers.}
Our ablation study, which created hybrid models by swapping components with baselines (Table~\ref{tab:abl_fg_po}), reveals the powerful synergy within UniAPO. While our feedback generator (FG) and prompt optimizer (PO) each provide significant, distinct benefits—mitigating visual token inflation and a lack of process-level supervision, respectively—all hybrid configurations underperform the complete UniAPO system. 

\setlength{\tabcolsep}{1.2pt}
\begin{table}[ht]
\small
\centering
\begin{tabular}{ccccc}
\toprule
\multirow{2}{*}{F-Mem} & \multirow{2}{*}{P-Mem} & Video CLS & \multicolumn{2}{c}{Video KE} \\ \cmidrule{3-5}
            &           & Occlusion layer      & Beauty             & Sport             \\ \midrule
Short       & Short     & 63.2 & 68.3 & 70.5                  \\
Short-long  & Short     & 66.7 & 71.3        &  75.6      \\
Short       & Short-long& 65.2&  70.9      &    74.0   \\
Short-long  & Short-long&\textbf{70.3}  & \textbf{74.7}  & \textbf{78.3}      \\  \bottomrule
\end{tabular}
\caption{Ablation of Short-Term and Long-Short Term memory mechanism in Feedback Memory (F-Mem) and Prompt Memory (P-Mem).}
\label{tab:abl_mem}
\end{table}

\subsubsection{Effect of each component in Memory Mechanism.}
Our ablation study confirms that UniAPO's dual memory system is critical. The long-term memory in Feedback Generation (FG) is essential for mitigating visual token inflation, while the long-term memory in Prompt Optimization (PO) provides process-level supervision. Removing either component cripples the system by introducing low-quality feedback or sub-optimal prompt, respectively. UniAPO's state-of-the-art performance is attributable to the synergy of these mechanisms in solving these core multimodal challenges.

\subsection{Case Study}
A case study on sport keyword extraction as shown in the Appendix reveals how UniAPO transforms a simple prompt into a sophisticated, hundred-line directive. This iterative evolution is driven by specific, class-level feedback—a product of our memory mechanism that successfully mitigates visual token inflation. The process history also confirms that the initial prompt's structure, even when simple, is critical for establishing a directed optimization path, highlighting the synergy of our approach.

\section{Conclusion}

We present \textbf{UniAPO}, the first unified framework for automated prompt optimization (APO) that operates effectively across text, image, and video tasks. By decoupling feedback modeling from prompt refinement through an EM-inspired scheme and introducing a long-short term memory mechanism, UniAPO overcomes key challenges in multimodal APO. Experiments show that UniAPO consistently surpasses existing baselines in both performance and generalization. We believe our approach paves the way for more robust and scalable prompt optimization in future multimodal language models.


\appendix
\section{Appendix}
\subsection{Details of Experimental Setting}
\subsubsection{Datasets}
For all text-based datasets, we adopt the data partitioning scheme from ERM~\cite{ERM}. Since the original Meme dataset lacks official validation and test splits, we partition it into training, validation, and test sets using a 3:3:4 ratio. The data splits for our in-house video datasets, designed for classification and keyword extraction tasks, are detailed in Table~\ref{tab:dataset_splits}. The splits for the classification tasks are designed to robustly evaluate the generalization performance of our models.

\begin{table}[ht]
\centering
\begin{tabular}{ccccc}
\toprule
\textbf{Task} & \textbf{Sub-task} & \textbf{Train} & \textbf{Validation} & \textbf{Test} \\
\midrule
\multirow{3}{*}{Text CLS}
 & BBH & 38 & 58 & 144 \\
  & Ethos & 798 & 200 & 200 \\
  & Liar & 3681 & 461 & 461 \\
\midrule
\multirow{1}{*}{Text GEN}
 & WebNLG & 200 & 300 & 300 \\
\midrule
\multirow{1}{*}{Image CLS}
 & Meme & 207 & 207 & 698 \\
\midrule
\multirow{2}{*}{Video CLS} & Static CLS & 100 & 100 & 834 \\
 & Occlusion & 91 & 91 & 204 \\
\midrule
\multirow{4}{*}{Video KE} & Beauty & 24 & 25 & 25 \\
 & Sport & 44 & 45 & 45 \\
 & Travel & 22 & 23 & 23 \\
 & Food & 22 & 22 & 22 \\
\bottomrule
\end{tabular}
\caption{Data splits for our in-house video datasets. The numbers represent the sample counts for each set. ``CLS'' denotes the classification task, ``GEN'' denotes the generation task and ``KE'' denotes the keyword extraction task.}
\label{tab:dataset_splits}
\end{table}

For our in-house video dataset, we process each video, which is approximately 1 to 2 minutes in duration, by uniformly sampling 8 frames to represent its visual content.
In addition to this visual information, we provide a rich set of accompanying textual modalities.
This includes the video's title, text extracted from stickers, text obtained via Optical Character Recognition (OCR) from the video frames, and the audio transcript generated by Automatic Speech Recognition (ASR).
\subsubsection{Evaluation Metrics.}
For the text-based tasks, we use the F1 score as our evaluation metric, following the evaluation methodology of ERM~\cite{ERM}. For the image classification task (Meme), we use the F1-micro score as our evaluation metric.
For the task of video binary classification, we use the F1 score as our evaluation metric.
For the task of video keyword extraction, we use the F1 score as our evaluation metric. Specifically, let $\tilde{y}$ denote the set of keywords predicted by the model for a given video, and let $y$ be the corresponding set of ground-truth keywords. We compute the cosine similarity between predicted keywords and ground-truth keywords using the BGE-m3~\cite{bge} model. A predicted keyword and a ground-truth keyword are considered a match if their similarity is greater than 0.9.

We then count the number of matched keywords. Let $\tilde{c}$ be the number of keywords in $\tilde{y}$ that match at least one keyword in $y$. Similarly, let $c$ be the number of keywords in $y$ that are matched by at least one keyword in $\tilde{y}$. Precision and recall are defined as follows:

\begin{equation}
    \text{Precision} = \frac{\tilde{c}}{|\tilde{y}|}, \quad \text{Recall} = \frac{c}{|y|}
    \label{eq:pr}
\end{equation}
where $|\cdot|$ denotes the cardinality of the set.

The F1 score for each sample is the harmonic mean of precision and recall:
\begin{equation}
    \text{F1} = \frac{2 \times \text{Precision} \times \text{Recall}}{\text{Precision} + \text{Recall}}
    \label{eq:f1}
\end{equation}
\subsubsection{Baselines.}
In the multimodal domains, where no established APO baselines exist, we extend EvolPrompt and ERM to multimodal settings, denoted as EvolPrompt* and ERM*, by adapting them to handle multimodal inputs. All methods are evaluated alongside naive prompting and CoT-style baselines for fair comparison.

\subsubsection{Implementation Details.}
We split each dataset into training, development, and test sets. In each iteration of prompt optimization, candidate prompts are trained on the training set, selected based on development performance, and evaluated on the test set.
We set the maximum number of training iterations to 12.
To prevent overfitting and reduce training time, we employ an early stopping strategy: training is terminated if the model's performance on the validation set does not show improvement.
For sequence generation tasks, we use a beam search with a beam size of 3, which is consistent with the setup in ERM~\cite{ERM}. We set the number of historical feedback instances and historical prompts used by UniAPO to 3 and 2, respectively.

All primary experiments use GPT-4o~\cite{achiam2023gpt} as the underlying MLLM across all stages of the UniAPO pipeline. Prompts are initialized with minimal handcrafted templates to simulate a low-resource setting, denoted as ``Simple Prompt''.
In additional experiments, we replace GPT-4o with QwenVL2.5-72B~\cite{bai2025qwen2} as the predictor to evaluate the generalization of UniAPO, while keeping the other components unchanged. We also explore settings with more structured initial prompts, as detailed in relevant sections.


\subsection{System Prompts of UniAPO}

\begin{tcolorbox}[breakable,
title=System Prompt of Predictor ($\mathcal{L}_{T}$),
colback=white,          
colframe=black,         
coltext=black,          
]
\begin{lstlisting}
Imagine you are a keyword extractor for short video ecosystem governance. I provide you the following video consisting of 8 frames, along with a title, stickers, ocr and asr. I hope you can determine whether this video meets the following policy. 
The title of the video is: {title}, the sticker texts of the video is: {stickers}, the ocr of the video is: {ocr}, the asr of the video is: {asr}, the frames of the video is: [VIDEO]
**POLICY**:
    <policy>{policy_str}</policy>
Please directly answer the extracted keywords list. The answer is wrapped with <answer> and </answer>.

**Output Format:
    <answer>[keyword1, keyword2, ...]</answer>
Answer:
\end{lstlisting}
\end{tcolorbox}

\begin{tcolorbox}[breakable,
title=System Prompt of the Cold Starting,
colback=white,          
colframe=black,         
coltext=black,          
]
\begin{lstlisting}
# Task Overview

You are tasked with creating a refined policy for a zero-shot keyword extraction model that handles challenging examples.

# Input Components

Original Prompt:

{user_prompt}

# Objective

Generate a detailed and robust POLICY that:

- Integrates All Details: Combines every element from the original prompt without oversimplifying or omitting critical instructions.
- Generate a structured policy in markdown format.

# Step-by-Step Reasoning & Verification Requirement

Before finalizing your answer, please perform the following using your internal chain-of-thought (which must not be visible in the final output):

- Understand the Task: Grasp the main objective, goals, requirements, constraints, and expected output.
- Minimal Changes: If an existing prompt is provided, improve it only if it's simple. For complex prompts, enhance clarity and add missing elements without altering the original structure.
- Clarity and Conciseness: Use clear, specific language. Avoid unnecessary instructions or bland statements.
- Preserve User Content: If the input task or prompt includes extensive guidelines or examples, preserve them entirely, or as closely as possible. If they are vague, consider breaking down into sub-steps. Keep any details, guidelines, examples, variables, or placeholders provided by the user.
- Constraints:
    - Confirm that the policy output adheres to the specified format with <policy> and </policy> tags.
    - *Do not add any output format.*
    - Do not add any input format and explanation of input content.
    - Do not add any examples.

**Note**: Only after thoroughly verifying your internal reasoning should you generate the final refined output.

# Output Format and Constraints
- Final Output Format:
    - The final refined policy must be wrapped within <policy> and </policy> tags.

- Word Limit:
    - The entire policy must not exceed 200 words.

# Final Output Template (Example):
<think>
... [Your short thinking process] ...
</think>
...
**Detailed Verified Refined Policy for Zero-Shot Keyword Extraction in Short Video Ecosystem Governance**:
<policy>
... [Your detailed and verified refined policy instructions go here] ...
</policy>

Output:
\end{lstlisting}
\end{tcolorbox}

\begin{tcolorbox}[breakable,
title=System Prompt of Feedback Generator ($\mathcal{L}_{F}$),
colback=white,          
colframe=black,         
coltext=black,          
]
\begin{lstlisting}
# Task Overview

You are tasked with creating a refined policy for a zero-shot keyword extraction model that handles challenging examples.

# Input Components

1. Original Prompt:

{user_prompt}

2. Additional Feedback (from problematic examples):

{feedback_str}

3. Observation:
The generated key cases based on the examples, which are provided under **INPUT**.

**INPUT**
[CASES_INPUT]

# Objective

Generate a detailed and robust POLICY that:

- Integrates All Details: Combines every element from the original prompt and the additional feedback without oversimplifying or omitting critical instructions.

# Step-by-Step Reasoning & Verification Requirement

Before finalizing your answer, please perform the following using your internal chain-of-thought (which must not be visible in the final output):

1. Break Down Each Requirement:

- Verify that the integration of both the original prompt and the additional feedback is complete.
- Confirm that clear instructions on handling each data field (video, predict_label, label) are provided.
- Ensure guidance for managing erroneous or missing keywords is explicitly addressed.

2. Cross-Check for Completeness:

- Ensure that no critical details are missing.
- Verify that the policy output will remain under the 4096-word limit.
- Confirm that the policy output adheres to the specified format with <policy> and </policy> tags.
- Do not add output format in the policy output.
- Do not lossing any other information in the policy output and shorten the policy output.
- Do not directly add the content of the **INPUT** in the policy output.
- Do not add the content of about **Continuous Improvement** in the policy output.

**Note**: Only after thoroughly verifying your internal reasoning should you generate the final refined output.

# Output Format and Constraints
- Final Output Format:
    - The final refined policy must be wrapped within <policy> and </policy> tags.

- Word Limit:
    - The entire policy must not exceed 4096 words.

- Clarity and Robustness:
    - Do not produce an oversimplified version. Every critical detail and instruction must be integrated to ensure the model can reliably handle difficult or noisy examples.

- Do not directly add the content of the **INPUT** in the policy output.

# Final Output Template (Example):
<think>
... [Your short thinking process] ...
</think>
...
**Detailed Verified Refined Policy for Zero-Shot Keyword Extraction in Short Video Ecosystem Governance**:
<policy>
... [Your detailed and verified refined policy instructions go here] ...
</policy>

Output:
\end{lstlisting}
\end{tcolorbox}

\begin{tcolorbox}[breakable,
title=System Prompt of the Prompt Evoluter ($\mathcal{L}_{E}$),
colback=white,          
colframe=black,         
coltext=black,          
]
\begin{lstlisting}
You must turn several policy documents into one consistent policy that keeps everything important, removes redundancies, and resolves contradictions.

INPUTS
 **Policies**: Multi-line string that contain **all** source policies (two or more).
   Make sure each policy is clearly delimited in the input (e.g. "### Policy 1", "### Policy 2", etc.).  
 **Restriction**: Optional extra constraints that the final policy must obey.

TASKS: PERFORM IN ORDER

1. **Extract rules.**  
   Read every source policy and mentally list all of its rules (no output yet).

2. **Merge identical / near-identical rules.**  
   a. If multiple policies express the same idea, keep the clearest wording.  
   b. Include the merged rule only once in the final list.

3. **Resolve conflicts.**  
   a. When two or more rules clash, decide which is preferable using general practicality and broad applicability.  
   b. Keep the chosen rule and discard the others.  
   c. Immediately after the kept rule, add a short bracketed note explaining why it was chosen, e.g.  
      "(Preferred over Policy 3 & 2 because it applies company-wide.)" 
   d. If conflicting rules can coexist under different conditions, keep all and state the conditions explicitly.

4. **Evaluate unique rules.**  
   a. For rules that appear in only one policy, keep them **only if** they are sensible, broadly useful, and not unreasonably specific.  
   b. If kept, rewrite for clarity or generality as needed.  
   c. If removed, record the reason in the "unique content" subsection of the Reason block wrapped with <think> and </think>.

5. **Quality check.**  
   Ensure the resulting policy:  
   - is logically consistent;  
   - covers every major scenario found in the inputs;  
   - contains no redundancy;  
   - satisfies {addi_restriction};  
   - is under 4096 words (including tags).

FORMATTING RULES

- The the content of Reason block is wrapped with <think> and </think>.
- The Detailed Merged Policy: is wrapped with `<policy>` and `</policy>`.
- Keep every tag (`<think>`, `</think>`, `<policy>`, `</policy>`) exactly as shown.  
- Do NOT nest tags or add other commentary outside the prescribed blocks. 
- Inside `**Policy**` include only the content of final policy string.

OUTPUT FORMAT (return nothing else)


### Reason
<think>
similarities
- ...one bullet per merged rule...

conflicts
- ...one bullet per conflict handled, showing the original rules and the resolution rationale...

unique content
- ...one bullet per unique rule, marked "KEPT" (with rewritten text) or "REMOVED" (with reason)...
</think>

...Detailed Merged Policy:
### Merged Policy:
<policy>[...Detailed Merged Policy...]</policy>


INPUTS:
**Policies**: {policy_str}
**Restriction**: {addi_restriction}

Output:
\end{lstlisting}
\end{tcolorbox}

\begin{tcolorbox}[breakable,
title=Prompt of Cases for Feedback Generation,
colback=white,          
colframe=black,         
coltext=black,          
]
\begin{lstlisting}
The title of No.{idx} video is: {title}, the sticker texts of No.{idx} video is: {stickers}, the ocr of No.{idx} video is: {ocr}, the asr of No.{idx} video is: {asr}, the label of No.{idx} video is: {label}, the prediction of No.{idx} video is: {pred}, the frames of No.{idx} video is: 
\end{lstlisting}
\end{tcolorbox}






\subsection{Results of UniAPO in Sport Keyword Extraction}

\begin{tcolorbox}[breakable,
title=Initial Prompt,
colback=white,          
colframe=black,         
coltext=black,          
]
\begin{lstlisting}
Keywords about sports.
\end{lstlisting}
\end{tcolorbox}
\begin{tcolorbox}[breakable,
title=Input Prompt,
colback=white,          
colframe=black,         
coltext=black,          
]
\begin{lstlisting}
The keyword extractor must identify and extract concise, relevant, and contextually accurate keywords specifically related to sports from the provided video. The assessment must consider the following components:  

1. \textbf{Title}: Analyze the title for any explicit mentions or implications of sports-related terms or activities.  
2. **Sticker Texts**: Evaluate text present in stickers within the video for sports-specific language or references.  
3. **OCR (Optical Character Recognition)**: Examine all visual text extracted from the video frames for keywords associated with sports, including sports names, terminologies, or related terms.  
4. **ASR (Automatic Speech Recognition)**: Process the spoken content within the video to identify sports-related mentions or themes.  
5. **Video Frames**: Contextually interpret visual elements in the frames (e.g., players, equipment, locations) to verify relevance to sports.  

The extracted keywords must be:  
- **Specific**: Pertinent to sports (e.g., "soccer," "basketball").  
- **Concise**: Avoiding redundancy or overly broad terms.  
- **Contextually Relevant**: Reflecting the actual theme or content presented in the video.  

Exclude generic or unrelated terms that do not directly relate to sports. Perform a holistic evaluation of all input components to ensure accurate keyword extraction. 
\end{lstlisting}
\end{tcolorbox}
\begin{tcolorbox}[breakable,
title=Generated Feedback,
colback=white,          
colframe=black,         
coltext=black,          
]
\begin{lstlisting}
**Regulation 1: Contextual Integration of Visual and Audio Cues}**

Develop a mechanism to integrate visual cues from the frames with audio cues from the ASR. If the frames depict a specific sport or activity and the ASR mentions related terms (e.g., "pass the ball" in a soccer scene), the system should prioritize extracting keywords that combine both sources of information. For example, in Video No.2, where soccer visuals and "pass the ball" audio cues were present, the keywords should include "Soccer Pass" or "Soccer Interaction." This ensures a holistic evaluation of the video content.


**Regulation 2: Refine extraction from video information by leveraging implicit contextual clues.**

- The extractor must analyze the video information (OCR and ASR data) to identify implicit references to sports activities or themes. For instance, if OCR data mentions "Naval Cadet Entrance Exam" and the video frames show individuals running, keywords like "1,000-meter Run" and "Naval Cadet Physical Exam" should be extracted to reflect the specific context.  
- When OCR or ASR data contains ambiguous or fragmented text, the extractor must cross-reference with video frames to clarify the context and extract relevant keywords. For example, if OCR mentions "LED Ball" and the video frames show a glowing basketball, the keyword "LED Basketball" should be included.  
- Avoid extracting keywords that are unrelated to the sports theme, even if they appear in OCR or ASR data. For example, terms like "celebration" or "product" should only be included if they are directly tied to the sports context presented in the video.  
- This regulation ensures that keywords derived from video information are contextually accurate and aligned with the visual content.


**Regulation 3: Frame-by-Frame Contextual Analysis**

Implement a systematic frame-by-frame analysis to identify recurring visual elements and actions. Each frame should be evaluated independently, and recurring sports-related elements (e.g., players, equipment, settings) should be flagged and cross-referenced across all frames. For example, if multiple frames show individuals skateboarding in a skatepark, keywords like "Casual Skateboarding Footage" or "Skatepark Activity" should be extracted to capture the broader context. This approach ensures that implicit and repeated sports activities are not overlooked.
\end{lstlisting}
\end{tcolorbox}
\begin{tcolorbox}[breakable,
title=Optimized Prompt,
colback=white,          
colframe=black,         
coltext=black,          
]
\begin{lstlisting}
### **Zero-Shot Sports-Related Keyword Extraction Policy**

As a keyword extractor for short video ecosystem governance, your role is to accurately and comprehensively identify sports-related keywords. Analyze all provided video components, including the title, stickers, OCR, ASR, and video frames, using the detailed guidelines below to ensure precise, contextually relevant, and specific keyword extraction.

---

### **Data Component Analysis**

1. **Title**:
   - Extract sports-related keywords directly or infer them based on context.
   - Prioritize specific sports disciplines, events, or activities.
   - Avoid extracting unrelated or generic terms unless tied explicitly to a sports context.

2. **Stickers**:
   - Analyze sticker texts for sports themes or terminology.
   - Extract only relevant sports-related stickers, excluding unrelated content.
   - Cross-reference sticker data with visual cues to confirm relevance.

3. **OCR (Optical Character Recognition)**:
   - Identify visible text in video frames for sports-related terms, brand names, team names, locations, or events.
   - Validate OCR keywords using visual evidence (e.g., equipment, uniforms).
   - Avoid speculative terms; derive only from explicit textual or visual evidence.

4. **ASR (Automatic Speech Recognition)**:
   - Extract spoken words referring to sports activities, athlete names, teams, tournaments, or events.
   - Cross-check ASR data with visual content to ensure accuracy and avoid overgeneralization.
   - Exclude speculative terms unsupported by visual or textual evidence.

5. **Video Frames**:
   - Observe frames for sports-related actions, objects, or symbols to supplement textual data.
   - Identify specific sports activities, equipment, uniforms, or event settings.
   - Use sequential frame analysis to detect recurring patterns or implied activities (e.g., gameplay, events).
   - Infer implied keywords (e.g., "penalty shootout," "boxing match") based on visual cues and frame progression.

---

### **Key Extraction Criteria**

1. **Specificity**:
   - Extract concise, meaningful, and specific keywords (e.g., "soccer," "basketball," "Olympics").
   - Avoid overly broad terms unless explicitly tied to a sports context.
   - Use detailed descriptors when supported by evidence (e.g., "100-meter dash" instead of "running").

2. **Relevance**:
   - Extract only sports-related keywords. Exclude irrelevant or overly broad terms.
   - Ensure keywords reflect the core theme or activity depicted in the video.

3. **Implied Keywords**:
   - Infer implied keywords when strong visual and contextual evidence supports their inclusion.
   - Example: If frames depict "Mbappe" and "Haaland" in a competitive setting, infer "Mbappe vs Haaland" or "penalty shootout."
   - Avoid speculative terms; derive implied keywords from visual and contextual cues.

4. **Contextual Integration**:
   - Cross-reference textual data (Title, Stickers, OCR, ASR) with visual cues from Video Frames to validate or enhance extracted keywords.
   - Example: If frames show a basketball court and ASR mentions "basketball," prioritize extracting "basketball game" or "basketball match."
   - Ensure extracted keywords align with the video's overarching theme.

5. **Prioritization of Visual Cues**:
   - Use visual evidence (e.g., equipment, player actions, event banners) to extract contextually relevant keywords.
   - Example: Frames showing athletes with a basketball should lead to the keyword "basketball," even if textual data is ambiguous.
   - Leverage sequential frame analysis to detect recurring patterns or implied themes (e.g., continuous gameplay).

6. **Avoiding Duplicates**:
   - Consolidate extracted keywords to avoid duplicates unless contextually significant.
   - Example: Use "soccer match" rather than repeating "soccer" and "match" separately.

7. **Error Handling and Missing Data**:
   - If components (e.g., Stickers, OCR, ASR) are absent or uninformative, rely on visual analysis (e.g., video frames).
   - Example 1: If frames show a skatepark and individuals performing tricks, extract "skateboarding," "skatepark," or "skateboarding tricks."
   - Example 2: If frames depict a racing track with vehicles, infer "racing event" or "amateur racing."
   - Avoid speculative keywords; derive terms from explicit or strongly implied evidence.

---

### **Advanced Techniques for Challenging Scenarios**

1. **Sequential Frame Analysis**:
   - Analyze the progression of frames to identify recurring patterns or implied activities.
   - Example 1: If consecutive frames depict players passing a soccer ball, infer "soccer passing" or "amateur soccer play."
   - Example 2: If frames show repeated actions in a boxing ring, infer "boxing match" or "amateur boxing."
   - Use frame progression to validate implied keywords (e.g., "penalty shootout").

2. **Enhanced ASR Integration**:
   - Cross-validate ASR data with visual evidence to ensure accuracy.
   - Example: If ASR mentions "Ronaldinho" and frames depict a soccer match, extract "Ronaldinho," "soccer," and "soccer match."
   - Avoid speculative terms unsupported by visual evidence.

3. **Contextual Integration of Visual and Textual Cues**:
   - Combine visual and textual data to infer detailed, contextually relevant keywords.
   - Example: If frames depict a billiards table with players engaged, infer "billiards skills" or "cue sports."

4. **Handling Ambiguous or Noisy Data**:
   - When OCR or ASR data is missing or uninformative, rely primarily on visual cues.
   - Example: If frames depict a boxing ring with athletes, extract "boxing match" and "amateur boxing."
   - Avoid overgeneralization by focusing on unique visual identifiers (e.g., team names, player numbers).

5. **Prioritize Frame-Based Contextual Evidence**:
   - Example: If frames depict a dirt track with sprint cars and stickers mention "URC Sprints," infer "sprint car racing" and "amateur racing."

6. **Avoid Overgeneralization**:
   - Focus on unique visual identifiers (e.g., equipment, player actions, event settings).
   - Example: Extract "goblet squat" rather than broad terms like "glutes workout" unless explicitly supported.

---

### **Output Guidelines**

- Submit extracted keywords as a comma-separated list enclosed within `<answer>` and `</answer>` tags.
- Ensure the keyword list is directly derived from video components by integrating textual and visual data seamlessly.
- Avoid unnecessary commentary or explanations in the output.

---

By systematically analyzing textual and visual data, prioritizing specificity and relevance, and addressing challenging examples with advanced integration techniques, this refined policy ensures accurate and reliable keyword extraction for short video ecosystem governance.
\end{lstlisting}
\end{tcolorbox}

\subsection{Results of EvolPrompt* in Sport Keyword Extraction}

\begin{tcolorbox}[breakable,
title=Initial Prompt,
colback=white,          
colframe=black,         
coltext=black,          
]
\begin{lstlisting}
Keywords about sports.
\end{lstlisting}
\end{tcolorbox}
\begin{tcolorbox}[breakable,
title=Optimized Prompt,
colback=white,          
colframe=black,         
coltext=black,          
]
\begin{lstlisting}
Identifying, pinpointing, and compiling key and essential terms related to sports.
\end{lstlisting}
\end{tcolorbox}

\subsection{Results of ERM* in Sport Keyword Extraction}

\begin{tcolorbox}[breakable,
title=Initial Prompt,
colback=white,          
colframe=black,         
coltext=black,          
]
\begin{lstlisting}
Keywords about sports.
\end{lstlisting}
\end{tcolorbox}
\begin{tcolorbox}[breakable,
title=Generated Feedback,
colback=white,          
colframe=black,         
coltext=black,          
]
\begin{lstlisting}
*Regulation #1: Enhance Specificity in Keyword Extraction for Sports Types and Content Types**

- When extracting keywords related to sports types, ensure granularity by prioritizing specific terms over general ones. For instance, instead of "Shooting Sports Highlights," extract precise terms such as "Air Pistol Shooting" or "Turkish Athlete" when identifiable in the video frames or ASR.

- For content types, include subcategories or contextual elements (e.g., "celebration," "stadium") that are visually or narratively significant within the video. Use visual cues from the video frames (e.g., player actions, equipment) and audio descriptions to refine keyword selection.\n- Cross-reference extracted keywords with the ASR and video frames to ensure alignment with the video\'s context and avoid overgeneralization.


**Regulation #2: Prioritize Contextual Relevance Derived from Video Frames**

- Enhance the extraction process by explicitly requiring the integration of visual cues from video frames. For example:

- Identify specific actions, objects, or environments depicted in the frames (e.g., "dirt track racing" in Example 1, "billiard table" in Example 2).

- Translate these visual elements into precise, contextually relevant keywords that align with the policy categories (e.g., "Fan Support" in Example 3).

- Implement a rule that prioritizes specific and actionable keywords over broad or generic terms. For instance, instead of extracting "sports humor," focus on extracting specific actions like "pass" or "challenge" as seen in Example 4.


**Regulation #3: Enforce Contextual Keyword Combination for Precision**

- Keywords must be contextually combined to capture the full meaning of the activity or event depicted in the video. This includes:

- Combining location and event elements (e.g., "Naval Cadet Entrance Exam" with "1,000-meter Run").
    - Merging comparative or descriptive elements with the activity (e.g., "Basketball Prototype vs Product").
    - Including relevant descriptors such as "amateur," "professional," or "casual" only when they are clearly supported by the video content.
    - Avoid overgeneralization by ensuring that extracted keywords are directly tied to the specific sports type or context shown in the video. For example, avoid generic terms like "amateur sports" when "amateur skateboarding" or "casual basketball" is more appropriate.
\end{lstlisting}
\end{tcolorbox}
\begin{tcolorbox}[breakable,
title=Optimized Prompt,
colback=white,          
colframe=black,         
coltext=black,          
]
\begin{lstlisting}
### **1. Objective**
This policy provides a robust framework for extracting precise, contextually relevant, and comprehensive keywords from short video content. It integrates visual and textual data to accurately represent the video's activities, objects, themes, and events. The goal is to ensure specificity, avoid redundancy, minimize errors, and handle challenging or ambiguous examples effectively.

---

### **2. Core Principles of Keyword Extraction**
- Extracted keywords must accurately reflect the video's central themes, dynamic actions, objects, and events as depicted in video frames and supported by textual data (ASR, OCR, stickers, and title).
- Emphasize specificity by prioritizing detailed, actionable, and contextually relevant keywords. Use generic terms (e.g., "sports") only as a fallback when specificity is not possible.
- Consolidate similar or overlapping keywords into precise terms unless distinct variations are explicitly emphasized in the video's content.

---

### **3. Data Integration and Analysis**
#### **3.1. Video Frames**
- **Primary Source**: Use video frames as the primary source for identifying specific actions, objects, and events. Prioritize dynamic actions (e.g., "goal scoring," "kickflip") and central objects (e.g., "boxing ring," "billiards table") visible across multiple frames.
- **Specificity in Actions**: Replace generic terms with specific descriptions. For example:
  - Replace "sports" with "horse racing" if horses and a racetrack are depicted.
  - Replace "fitness" with "front dumbbell raises" if the video shows this specific exercise.
- **Temporal Progression**: Consider sequences of actions across frames to derive comprehensive keywords. For example, if the video shows a goal being scored followed by a celebration, include both "goal scoring" and "celebration."
- **Consistency**: Cross-reference keywords across all frames to ensure consistency in representing the video's core theme. For example, if a boxing ring is visible in all frames, ensure "boxing ring" is included.

#### **3.2. ASR (Audio Speech Recognition)**
- **Complement Visual Analysis**: Use ASR data to identify spoken phrases that add context or confirm visual findings. For example:
  - If the ASR mentions "kickflip" during a skateboarding video, ensure "kickflip" is included as a keyword.
- **Prioritize Relevance**: Extract keywords from ASR phrases that directly describe actions, events, or objects relevant to the video frames. Disregard irrelevant or noisy ASR outputs unless supported by visual or other textual data.

#### **3.3. OCR (Optical Character Recognition)**
- **Contextual Integration**: Extract keywords from text visible in the video, such as banners, signs, or equipment labels. For example:
  - If OCR identifies "URC SPRINTS," include "sprint cars" or "dirt track racing."
  - If OCR reads "Pool Tournament 2023," include "pool" and "billiards."
- **Cross-Validation**: Validate OCR findings against visual and ASR data to ensure consistency and accuracy. Disregard misleading or irrelevant OCR outputs unless corroborated by other data sources.

#### **3.4. Stickers and Title**
- **Stickers**: Use sticker text to provide additional context. For example:
  - If a sticker reads "Game On!" and the video depicts a basketball court, include "basketball" or "game" as keywords.
- **Title**: Use the title as a guide for overall context but ensure alignment with observable actions, objects, or events in the video frames and textual data.

---

### **4. Specificity, Coverage, and Redundancy**
#### **4.1. Enhancing Specificity**
- Use detailed and actionable keywords rather than generic terms. For example:
  - Replace "sports" with "skateboarding" if skateboarding is depicted.
  - Replace "fitness" with "deadlift workout" if the video shows a deadlift exercise.
- Consolidate overlapping keywords unless distinct variations are emphasized. For example:
  - Merge "Amateur Arm Wrestling" and "Arm Wrestling" into "Arm Wrestling" unless the amateur nature is central to the theme.

#### **4.2. Comprehensive Coverage**
- Analyze all input components (video frames, ASR, OCR, stickers, and title) to ensure no critical keywords are missed.
- Cross-reference extracted keywords across data sources to verify that they comprehensively represent the video's content.
- Derive context-specific keywords from ambiguous data based on the most likely interpretation of available information.

#### **4.3. Avoiding Redundancy**
- Avoid repetitive keywords unless they describe distinct aspects of the content. For example:
  - Include both "goal scoring" and "celebration" if these are separate actions depicted in the video.
  - Avoid duplicating "billiards" and "pool" unless both terms are contextually significant.

---

### **5. Error Handling and Edge Cases**
#### **5.1. Erroneous Data**
- Exclude irrelevant or misleading keywords caused by noisy ASR or OCR outputs unless supported by other contextual data. For example:
  - Ignore an OCR output of "SALE" if it is unrelated to the video's theme.
- Use visual data as the primary source to confirm or reject noisy textual inputs.

#### **5.2. Missing Data**
- If specific data fields (e.g., ASR, OCR) are unavailable, rely more heavily on the available sources while maintaining specificity and contextual relevance.
- If visual data lacks specificity, use textual data (ASR, OCR, stickers, and title) to infer likely themes or activities.

#### **5.3. Fallback Strategy**
- Use broad keywords (e.g., "sports") only when visual and textual data lack the specificity needed for detailed descriptions.

---

### **6. Quality Assurance**
#### **6.1. Iterative Testing**
- Conduct iterative testing across diverse and challenging examples to ensure adherence to the policy.
- Update the policy regularly based on observed errors and emerging edge cases to improve robustness.

#### **6.2. Final Review**
- Perform a final review to ensure:
  - All keywords are relevant, specific, and non-redundant.
  - Keywords comprehensively represent the video's content.
  - Justification exists for any broad or fallback keywords used.

---

### **7. Output Requirements**
- Present keywords as a concise, comma-separated list.
- Example format:
  - `<answer>[keyword1, keyword2, ...]</answer>`

---

### **8. Illustrative Examples**
#### **Example 1**:
- **Video Frames**: Show individuals performing "front dumbbell raises."
- **ASR**: Mentions "dumbbell workout."
- **OCR**: Reads "Strength Training."
- **Stickers**: Include "Fitness Goals."
- **Extracted Keywords**: `<answer>[front dumbbell raises, strength training, dumbbell workout]</answer>`

#### **Example 2**:
- **Video Frames**: Depict a skateboarding activity in a skatepark.
- **ASR**: Mentions "kickflip."
- **OCR**: Displays "Skateboarding Championship."
- **Stickers**: Say "Extreme Sports."
- **Extracted Keywords**: `<answer>[skateboarding, kickflip, skatepark]</answer>`

#### **Example 3**:
- **Video Frames**: Show a billiards table with players.
- **ASR**: Includes "Eight-ball, your turn."
- **OCR**: Reads "Pool Tournament 2023."
- **Stickers**: Include "Game Night."
- **Extracted Keywords**: `<answer>[billiards, pool, eight-ball]</answer>`


\end{lstlisting}
\end{tcolorbox}
\clearpage
\bibliography{aaai2026}
\end{document}